\documentclass[cameraready]{Interspeech}

\title{Contrastive Training with LLM-generated Near-Misses for Robust Code-Switching Speech Recognition}

\author[affiliation={1}, equalcontribution]{Tung}{X. Nguyen}
\author[affiliation={1}, equalcontribution]{Hieu Minh}{Truong}
\author[affiliation={1}]{Giang Son}{Nguyen}
\author[affiliation={1,2}]{Nhu}{Vo}
\author[affiliation={1,3}]{Wray}{Buntine}
\author[affiliation={1}]{\mbox{Dung~D.~Le}}{}

\address{
    $^1$ VinUniversity, Vietnam  \\
    $^2$ University of Technology Sydney, Australia \space
    $^3$ Monash University, Australia
}

\email{tung.nx@vinuni.edu.vn}
\email{\{tung.nx, hieu.tm2, son.ng, nhu.vd, wray.b, dung.ld\}@vinuni.edu.vn}


\usepackage{comment}
\usepackage{enumitem}
\usepackage{amsmath,amssymb}

\usepackage{CJKutf8}

\usepackage{multirow}

\usepackage[utf8]{inputenc}
\usepackage[T5]{fontenc}
\usepackage{algorithm}
\usepackage{algpseudocode}

\let\Lev\relax
\DeclareMathOperator{\Lev}{Lev}

\begin{document}

\maketitle

\begin{abstract}

Code-switching (CS), the alternation between multiple languages within a single utterance, remains challenging for Automatic Speech Recognition (ASR). To address this issue, we propose a Point-of-Interest (POI)-aware contrastive training framework that improves recognition at CS-critical regions. We first identify CS spans by adopting POI detection method from literature, then construct acoustically plausible \textit{near-miss} hypotheses by perturbing POIs in ASR $N$-best outputs and expanding candidates with a large language model. Hard but plausible negatives are retained through filtering with acoustic, phonemic, and textual constraints. Finally, we fine-tune Whisper-small with LoRA using a POI-weighted cross-entropy anchor objective together with a multi-negative contrastive ranking loss. Experiments on CS-FLEURS (\texttt{cmn-eng}) and ViMedCSS (\texttt{vie-eng}) show consistent reductions of over 2\% in both general and CS-aware error rates compared to standard LoRA fine-tuning.
\end{abstract}

\section{Introduction}
\label{sec:intro}

In Automatic Speech Recognition (ASR), \textbf{code-switching} (CS), the alternation between two or more languages within a single utterance or discourse, presents a unique challenge \cite{agro2025csasr_survey}. The presence of CS terms introduces language confusion and phonetic ambiguity, which can degrade the accuracy of the ASR decoder \cite{agro2025csasr_survey}. Empirically, the most severe recognition errors tend to cluster around these CS regions \cite{liu2025underlens,agro2025csasr_survey}.

While fine-tuning on (audio, transcript) pairs with code-switching reduces overall error rate across the entire utterance \cite{yan25c_interspeech, nguyen2026vimedcssvietnamesemedicalcodeswitching}, the effect of fine-tuning on robustness in CS regions remains understudied. Furthermore, standard fine-tuning objectives lack an explicit signal to target these confusable spans.

To enhance the accuracy of ASR models specifically at CS regions, we propose a contrastive learning objective that encourages the model to prefer the ground-truth CS terms over acoustically plausible but incorrect transcriptions (``\textit{near-misses}'', see Table~\ref{tab:qualitative_examples}). Concretely, we first collect the $N$-best hypotheses produced by the ASR model over the training data, then identify CS regions by adopting the Point-of-Interest (POI) detection method of \cite{ugan2025pier}. Next, we perturb only the POIs to construct acoustically plausible near-miss hypotheses and use an external large language model (LLM) to generate additional near-miss candidates. Finally, we optimize a maximum-likelihood objective on the anchor together with a multi-negative contrastive ranking loss inspired by preference-based alignment methods \cite{rafailov2023dpo,xu2024cpo}.

Our contributions are:
\begin{itemize}[leftmargin=*]
  \item \textbf{POI-local near-miss generation} (Section~\ref{sec:gen}). We introduce CS-NMG, an acoustic-aware pipeline that constructs \emph{POI-local} near-misses seeded from $N$-best hypotheses and optionally expanded offline by an LLM, while preserving audio plausibility.
  \item \textbf{Contrastive alignment for CS-ASR} (Section~\ref{sec:align}). We propose a training objective that combines a POI-weighted cross-entropy (CE) anchor with multi-negative contrastive ranking, explicitly separating common POI confusions rather than merely upweighting POI tokens.
  \item \textbf{Improvements over baselines on general and CS-critical evaluation} (Section~\ref{sec: exp}). Across two CS benchmarks (CS-FLEURS \texttt{cmn-eng} and ViMedCSS \texttt{vie-eng}), our approach improves over strong baselines on both traditional Word Error Rate (WER) and Point-of-Interest Error Rate (PIER).
\end{itemize}

\begin{table}[t]
\centering
\small
\renewcommand{\arraystretch}{1.2}
\begin{tabular}{@{} >{\bfseries}l p{0.75\columnwidth} @{}}
\toprule
\multicolumn{2}{@{}l}{\textbf{(1) Near-Miss Generation via LLM}} \\
\midrule
Reference & enzyme 5 alpha \textbf{\textcolor{blue}{reductase}} được tạo ra \\
ASR output & enzyme 5 alpha \textbf{\textcolor{red}{reduc tây giờ}} được tạo ra. \\
Near-miss 1 & enzyme 5 alpha \textbf{\textcolor{red}{ri đắc tê giờ}} được tạo ra. \\
Near-miss 2 & enzyme 5 alpha \textbf{\textcolor{red}{reduc tay giờ}} được tạo ra. \\
\multicolumn{2}{@{}p{0.95\columnwidth}@{}}{\footnotesize \textit{Insight:} The LLM-generated candidates preserve sentence structure while introducing phonetic confusions localized at the code-switching point-of-interest.} \\
\midrule
\midrule
\multicolumn{2}{@{}l}{\textbf{(2) Robustness via Contrastive Fine-Tuning}} \\
\midrule
Standard FT & enzyme 5 alpha \textbf{\textcolor{red}{reduc tây giờ}} được tạo ra \\
Contrastive FT & enzyme 5 alpha \textbf{\textcolor{blue}{reductase}} được tạo ra \\
\multicolumn{2}{@{}p{0.95\columnwidth}@{}}{\footnotesize \textit{Insight:} Contrastive fine-tuning suppresses acoustically plausible substitutions and correctly recovers the intended code-switching point-of-interest.} \\
\bottomrule
\end{tabular}

\caption{Qualitative example from ViMedCSS dataset \cite{nguyen2026vimedcssvietnamesemedicalcodeswitching}. (1) LLM near-miss candidates under acoustic and phonetic constraints. (2) Contrastive fine-tuning recovers the code-switched POI.}
\label{tab:qualitative_examples}
\vspace{-3mm}
\end{table}

\begin{figure*}[t]
    \centering
    \includegraphics[width=1.0\textwidth]{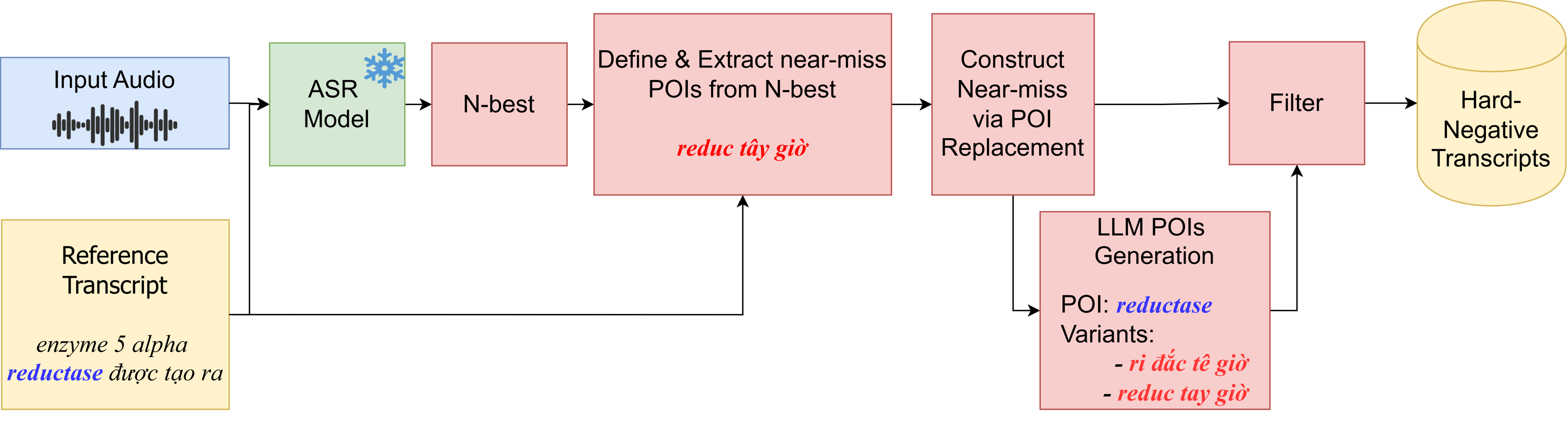}
    \caption{Overview of the proposed CS-NMG pipeline for code-switching near-miss generation.}
    \label{fig:pipeline}
\end{figure*}

\section{Related Work}
\label{sec:related}

\noindent \textbf{Code-switching ASR.}
CS-ASR remains challenging as errors concentrate on embedded-language words/entities and switch-boundary neighborhoods \cite{agro2025csasr_survey,liu2025underlens}. Many approaches inject language awareness such as language identification (LID) supervision to reduce confusion at switch points \cite{zeng19_interspeech,punjabi2020joint_asr_lid,li2019cs_ctc_lid,wang-li-2023-calcs}. For CS-critical evaluation, Point-of-Interest Error Rate (PIER) highlights errors produced by Point-of-Interest (POI) spans which is bundled with other error types in plain WER \cite{ugan2025pier}. Language-balance reweighting improves robustness on these POI regions \cite{ugan2025adaptingbalance}. General domain benchmarks such as CS-FLEURS \cite{yan25c_interspeech} and domain-focused resources like ViMedCSS \cite{nguyen2026vimedcssvietnamesemedicalcodeswitching} provide (audio, transcript) pairs where CS occurs. Benchmarking on these data sources reveals persistent difficulty in distinct-script and specialized-domain switching. Data-centric augmentation (e.g., code-switched text-to-speech or synthetic phrase-mixing) further mitigates transcript scarcity \cite{sharma20c_interspeech,nguyen2025norealcs}.

\noindent \textbf{Post-correction and LLM-based approaches.}
Post-decoding methods rescore or correct outputs using multiple hypotheses (e.g., $N$-best lists/lattices). Prior to LLMs, external LM fusion and discriminative rescoring were widely used \cite{sriram18_interspeech,tsunoo22_interspeech,ogawa19_interspeech}; more recently, instruction-tuned LLMs have been used to expand or rerank $N$-best hypotheses, including in code-mixed settings \cite{tur2024progres,kumar-akhtar-2025-clear}. These typically add inference-time post-processing, motivating training-time alternatives that keep ASR-only inference.

\noindent \textbf{Preference-guided alignment.}
Preference-based objectives optimize models from preferred vs.\ dispreferred outputs, including RLHF \cite{ouyang2022instructgpt}, DPO \cite{rafailov2023dpo}, and contrastive variants such as CPO \cite{xu2024cpo}. Applying such alignment specifically to POI-local CS confusions remains limited; we instantiate it for CS-ASR using acoustically grounded near-miss hypotheses.

\noindent \textbf{Discriminative training and $N$-best objectives.}
Sequence-level criteria optimize expected loss over competing hypotheses. MWER minimizes expected word errors using sampled or $N$-best hypotheses for seq2seq ASR \cite{prabhavalkar2018mwer}, and efficient $N$-best variants exist for RNN-T \cite{guo20_interspeech}. In contrast, we use the $N$-best neighborhood to \emph{seed} controlled \emph{POI-local} near-misses and train with contrastive ranking, rather than optimizing an expectation over arbitrary full-hypothesis edits.

\noindent \textbf{Positioning of our work.}
We differ from MWER-style training by constructing POI-local hard negatives (embedded spans and switch-boundary neighborhoods) and optimizing reference-vs-near-miss ranking. We also differ from LLM-based correction by using the LLM offline to expand POI replacements, keeping decoding standard ASR-only.

\section{Near-Miss Data Generation}
\label{sec:gen}
We propose CS-NMG, an acoustic-aware code-switching near-miss generation pipeline (Fig.~\ref{fig:pipeline}) that constructs hard-negative transcripts by perturbing only points-of-interest (POIs) while preserving plausibility under the input audio. 
CS-NMG is executed once \emph{offline} using a fixed seed ASR checkpoint (parameters frozen for decoding and teacher-forced scoring); the resulting near-miss pool is cached and reused across all training epochs.

\subsection{POI candidate pool from {\it N}-best and LLM generation}
Given an utterance (audio) $x$ with reference transcript $y^*$, we decode an $N$-best list using a fixed seed ASR model with parameters $\theta$:
\begin{equation}
\mathcal{H}(x)=\{h_i\}_{i=1}^{N}, \qquad s_i=\log p_\theta(h_i\mid x).
\label{eq:nbest}
\end{equation}
Here, $\mathcal{H}(x)$ is the seed ASR $N$-best set, $h_i$ is its $i$-th hypothesis, and $s_i$ is the seed-model log score. Following \cite{ugan2025pier}, let $E(y^*)$ denote the set of embedded-language spans in the reference $y^*$ (maximal contiguous segments whose language ID differs from the matrix language; detected as in Sec.~\ref{sec: exp}). For each span $e=(s_e,\dots,t_e)$, we define its switch-boundary neighborhood by expanding the boundaries by $\pm r$ tokens:
\begin{equation}
\mathrm{nbhd}(e;r) = [\max(1,s_e-r), \min(|y^*|,t_e+r)].
\end{equation}
The POI index set is $I(y^*)=\bigcup_{e\in E(y^*)}\mathrm{nbhd}(e;r)$ (duplicates merged).


Next, we query an LLM \emph{offline} to expand the POI candidate set for each $j$. The prompt provides the reference transcript $y^*$ with the target POI span $j$ marked, together with the raw $N$-best POI candidates; the LLM outputs a short list of additional replacement strings for the marked POI (no other edits). We merge the $N$-best and LLM suggestions and de-duplicate to obtain the final candidate set $\tilde{C}(j)$ for POI $j$.

\subsection{Near-miss construction by POI replacement}
For each POI $j\in I(y^*)$, we construct near-miss transcripts by replacing \emph{only} the POI span in $y^*$ with a sampled replacement $c\in\tilde{C}(j)$, leaving the rest of the transcript unchanged; we denote the resulting near-miss transcript by $\tilde{y}$. We sample up to $K$ near-misses per utterance, typically by replacing one POI at a time for efficiency.

\subsection{Filtering Gate}
We filter each near-miss $\tilde{y}$ to ensure acoustic plausibility and to control POI-level hardness. First, we apply an acoustic gate:
\begin{equation}
\log p_\theta(\tilde{y}\mid x)
\ \ge\
\max_{1\le i\le N}\log p_\theta(h_i\mid x) - \Delta .
\label{eq:acoustic_gate}
\end{equation}
Here, $\Delta$ is the acoustic-margin threshold. Let $c$ be the inserted POI segment at index $j$ and $y^*[j]$ the corresponding reference span. We then measure (i) how much $c$ differs in surface form and (ii) how close it remains in pronunciation:
\begin{align}
d_{\text{txt}}(c, y^*[j]) &=
\frac{\mathrm{Lev}(c,\, y^*[j])}{\max\{|c|,\ |y^*[j]|\}},
\label{eq:dtxt}
\\
d_{\text{ph}}(c, y^*[j]) &=
\frac{\mathrm{Lev}(\Phi(c),\, \Phi(y^*[j]))}{\max\{|\Phi(c)|,\ |\Phi(y^*[j])|\}},
\label{eq:dph}
\end{align}
where $\mathrm{Lev}$ measures Levenshtein distance \cite{Levenshtein1965BinaryCC} and $\Phi$ maps text to phonemes via a G2P model \cite{BISANI2008434}. We keep candidates satisfying:
\begin{equation}
d_{\text{txt}}(c, y^*[j]) \ge \tau_{\text{txt}}
\qquad
d_{\text{ph}}(c, y^*[j]) \le \tau_{\text{ph}}
\label{eq:poi_gates}
\end{equation}
where $\tau_{\text{txt}}$ enforces \emph{hardness} by requiring sufficient textual deviation and $\tau_{\text{ph}}$ enforces \emph{plausibility} with phonetic proximity.

\noindent In experiments, we instantiate five variants:
\emph{No filter} (no gating),
\emph{Acoustic} (Eq.~\ref{eq:acoustic_gate}),
\emph{Ac.+Text} (Eq.~\ref{eq:acoustic_gate} and $d_{\text{txt}}\negthinspace\ge\negthinspace\tau_{\text{txt}}$),
\emph{Ac.+Ph.} (Eq.~\ref{eq:acoustic_gate} and $d_{\text{ph}}\negthinspace\le\negthinspace\tau_{\text{ph}}$),
and \emph{Ac.+Ph.+Text} (our \textbf{tri-level} filter: Eq.~\ref{eq:acoustic_gate} plus both constraints in Eq.~\ref{eq:poi_gates}).

\section{Alignment Training Strategy}
\label{sec:align}
We fine-tune the ASR model using a maximum-likelihood (CE/WCE) anchor on the reference transcript and a contrastive ranking loss that prefers reference transcripts over acoustically plausible POI-local near-misses generated by CS-NMG (Sec.~\ref{sec:gen}).

\subsection{Cross-entropy anchor (CE / WCE)}
Following language-balance training for CS \cite{ugan2025adaptingbalance}, let $m_t\in\{0,1\}$ indicate whether target token $y_t^*$ lies within a POI span. We upweight POI tokens via
\begin{equation}
w_t = 1 + (\alpha_{\text{wce}}-1)m_t,
\label{eq:wt}
\end{equation}
and optimize the weighted cross-entropy (WCE)
\begin{equation}
\mathcal{L}_{\mathrm{WCE}}(x,y^*)
=
-\frac{1}{\sum_t w_t}\sum_{t} w_t \log p_\theta(y_t^*\mid x,y_{<t}^*).
\label{eq:wce}
\end{equation}
\noindent\textbf{CE vs.\ WCE.} Standard cross-entropy is Eq.~\ref{eq:wce} with $\alpha_{\text{wce}}{=}1$ (thus $w_t{=}1$ for all tokens). For WCE, we set $\alpha_{\text{wce}}{>}1$ and tune it on the dev set.

\subsection{Contrastive alignment with near-misses}
For any candidate transcript $y$, we use the length-normalized teacher-forced score
\begin{equation}
S_\theta(y;x)
=
\frac{1}{|y|}
\sum_{t=1}^{|y|} \log p_\theta(y_t\mid x,y_{<t}),
\label{eq:score}
\end{equation}
which avoids bias toward shorter candidates when POI-local insertions/deletions occur.

\noindent\textbf{Fixed negative selection.}
Given the CS-NMG near-miss pool $\tilde{\mathcal{Y}}(x,y^*)$, we select $K$ negatives via a fixed policy that (i) enforces the enabled CS-NMG gates and (ii) promotes diversity across POI categories (embedded / boundary) and POI edit type (substitution/insertion/deletion):

\noindent We then optimize an InfoNCE-style objective \cite{oord2018cpc,gutmann2010nce} that ranks $y^*$ above the selected negatives $\{y_k^-\}_{k=1}^{K}$:
\begin{equation}
\mathcal{L}_{\mathrm{CL}}(x)
=
-\log
\frac{\exp(\beta\,S_\theta(y^*;x))}
{\exp(\beta\,S_\theta(y^*;x))+\sum_{k=1}^{K}\exp(\beta\,S_\theta(y_k^-;x))},
\label{eq:infonce}
\end{equation}
with $\beta{=}1$ in all experiments. The final training loss is
\begin{equation}
\mathcal{L}
=
\mathcal{L}_{\mathrm{WCE}}
+\lambda_{\mathrm{CL}}\,\mathcal{L}_{\mathrm{CL}}.
\label{eq:loss_nce}
\end{equation}

\section{Experiments}
\label{sec: exp}
\subsection{Setup and Datasets}

\begin{table}[t]
\centering
\footnotesize 
\setlength{\tabcolsep}{3pt} 
\caption{Main results on \texttt{cmn-eng} (CS-FLEURS) and \texttt{vie-eng} (ViMedCSS). We compare cross-entropy baselines (CE/WCE), a sequence-level baseline (MWER), and our contrastive training with near-miss (NM) negatives. The full model (WCE+CL with tri-level filtering) achieves the lowest WER and PIER on both datasets. Lower is better ($\downarrow$).}
\label{tab:main_results}
\resizebox{\columnwidth}{!}{
\begin{tabular}{l | c c | c c}
\toprule
\multirow{2}{*}{\textbf{Method}} & \multicolumn{2}{c|}{\texttt{cmn-eng}} & \multicolumn{2}{c}{\texttt{vie-eng}} \\
& \textbf{WER} & \textbf{PIER} & \textbf{WER} & \textbf{PIER} \\
\midrule

\multicolumn{5}{l}{\textit{Baselines}} \\
\quad CE & 16.67 & 17.25 & 24.72 & 21.95 \\
\quad WCE & 16.42 & 16.68 & 24.21 & 21.18  \\
\quad MWER & 15.75 & 16.41 & 23.82 & 20.84 \\

\midrule

\multicolumn{5}{l}{\textit{Ours}} \\
\quad CE + CL ($N$-best NM) & 15.64 & 16.21 & 23.16  & 20.11 \\
\quad WCE + CL ($N$-best NM)& 14.93 & 15.72  & 22.86  & 19.10 \\
\quad \textbf{WCE + CL (tri-level)} & \textbf{14.06} & \textbf{15.10} & \textbf{21.87} & \textbf{18.74} \\

\bottomrule
\end{tabular}%
}
\end{table}

\noindent \textbf{Datasets and Metrics.}
We evaluate on two code-switching ASR benchmarks: (i) Mandarin--English (\texttt{cmn-eng}) from CS-FLEURS \cite{yan25c_interspeech} and (ii) Vietnamese--English (\texttt{vie-eng}) from ViMedCSS \cite{nguyen2026vimedcssvietnamesemedicalcodeswitching}, a domain-specific (medical) benchmark.
For CS-FLEURS, we fine-tune on the \textsc{CS-FLEURS-XTTS} training split and evaluate on the human-read \textsc{CS-FLEURS-READ} test split for the same pair (Mandarin matrix with English insertions).
For ViMedCSS, we fine-tune on the Train split and evaluate on the Test split, additionally reporting the Hard split when analyzing rare medical terminology.
We report WER together with POI-focused PIER \cite{ugan2025pier}. To compute PIER, we align each hypothesis $\hat{y}$ to the reference $y^*$ (word-level for whitespace-tokenized languages; character-level for CS-FLEURS) and measure the normalized Levenshtein distance restricted to POI positions:
\begin{equation}
\mathrm{PIER}(y^*,\hat{y})=
\frac{\Lev\negthinspace \left(
       y^* [I(y^*)],\,\hat{y}[I(y^*)]
       \right) }{
\left| 
  y^* [I(y^*)] 
  \right| 
  } .
\label{eq:pier}
\end{equation}


\noindent \textbf{Generation Configuration:}
We follow PIER \cite{ugan2025pier} to tag POIs. For \texttt{cmn-eng}, English POIs are detected via Latin-script tokens; for \texttt{vie-eng}, we apply token-level language ID on whitespace-tokenized words (excluding punctuation-only and numeric tokens).
Near-miss POI replacements are generated by querying Gemini 2.5 Pro via API \cite{comanici2025gemini25pushingfrontier} with temperature $=1$, top-p $=0.95$, single-candidate output, where the prompt provides the full reference transcript $y^*$ with the POI span marked and includes the raw $N$-best POI pool $\tilde{C}_{\text{$N$-best}}(j)$ as grounded candidate hints.
The LLM outputs only a short list of POI replacement strings (no other edits); we de-duplicate and apply the same CS-NMG filtering gates (Sec.~\ref{sec:gen}) before training. After POI replacement, we apply the optional filtering configurations described in Sec.~\ref{sec:gen} (acoustic gate, and optional text/phoneme constraints). Unless stated otherwise, we choose $\tau_{\text{ph}}=0.6$ and $\tau_{\text{txt}}=0.4$ for POI-local filtering by grid search. For phonetic normalization, we use deterministic toolkits: Mandarin tokens are converted to tone-marked Pinyin via \texttt{pypinyin} \cite{pypinyin2025}, English tokens to ARPAbet via \texttt{g2p\_en} \cite{g2pE2019}, and Vietnamese tokens to tone-preserving syllable units via \texttt{underthesea} \cite{underthesea2017}. All distances are computed as normalized Levenshtein distances over these phonetic sequences \cite{Levenshtein1965BinaryCC}.

\noindent \textbf{Backbone and Training Protocol:}
We adopt Whisper-small \cite{radford2022robustspeechrecognitionlargescale} as the backbone and fine-tune using LoRA, following prior Whisper adaptation work \cite{song24_interspeech,xu24h_interspeech}. For LoRA, we set the rank to $r=16$, scaling factor $\alpha=32$, dropout to $0.05$, and learning rate to 0.001. We decode with beam search and set $N=10$ to construct the $N$-best list for near-miss generation. Based on development experiments, we fix the number of sampled near-misses to $K=5$ per utterance and set the contrastive weight to $\lambda_{\text{CL}}=0.1$ (larger values yielded diminishing returns).
The acoustic margin is fixed at $\Delta=4.0$. For likelihood baselines, we report both CE and language-balance WCE; the WCE scaling $\alpha_{\text{wce}}$ is tuned on dev and set to $1.7$ for CS-FLEURS and $2.0$ for ViMedCSS. All other hyperparameters are kept fixed across experiments. For contrastive training, all candidate scores use the length-normalized sequence score in Eq.~\ref{eq:score}. For the baselines, we also compare with MWER approach using the same setting as \cite{prabhavalkar2018mwer}.

\subsection{Results and Analysis}

\begin{table}[t]
\centering
\footnotesize
\setlength{\tabcolsep}{3.5pt}
\caption{Attribution ablations on \texttt{cmn-eng} (CS-FLEURS) and \texttt{vie-eng} (ViMedCSS). We compare near-miss (NM) sources (\textit{$N$-best} vs.\ \textit{$N$-best+LLM} under \textit{No filter}) and gating strategies (applied to \textit{$N$-best+LLM}) for selecting hard-but-plausible NMs. \textbf{ \#NM/utt} is the conditional mean of selected NMs over utterances retaining at least one selected near-miss under each gate. The full tri-level gate \textit{Ac.+Ph.+Text} performs best overall, except for cmn--eng PIER. Lower is better ($\downarrow$).}
\label{tab:ablations}
\resizebox{\columnwidth}{!}{
\begin{tabular}{l | c c c | c c c}
\toprule
\multirow{2}{*}{\textbf{Variant}} 
& \multicolumn{3}{c|}{\texttt{cmn-eng}} 
& \multicolumn{3}{c}{\texttt{vie-eng}} \\
& \textbf{WER} & \textbf{PIER} & \textbf{\#NM/utt}& \textbf{WER} & \textbf{PIER} & \textbf{\#NM/utt} \\
\midrule

\multicolumn{7}{l}{\textit{No filter}} \\
\quad $N$-best 
& 14.93 & 15.72 & 1.40
& 22.86 & 19.10 & 1.22 \\
\quad $N$-best + LLM 
& 15.06 & 15.28 & 6.00
& 22.19 & 19.65 & 6.00 \\

\midrule
\multicolumn{7}{l}{\textit{Filter Gate}} \\
\quad Ac. only 
& 15.55 & 15.18 & 5.75
& 24.03 & 19.56 & 4.93 \\
\quad Ac. + Ph. 
& 14.50 & 15.16 & 5.65
& 23.17 & 19.73 & 4.92 \\
\quad Ac. + Text 
& 14.12 & \textbf{14.69} & 3.76
& 24.03 & 19.71 & 3.82 \\
\quad \textbf{Ac.+Ph.+Text} 
& \textbf{14.06} & 15.10 & 3.77
& \textbf{21.87} & \textbf{18.74} & 3.81 \\

\bottomrule
\end{tabular}
}
\end{table}

Table~\ref{tab:main_results} compares WER and PIER across different training strategies.

\textbf{Baselines.} 
Plain CE provides a strong LoRA baseline, but leaves substantial residual errors on CS-critical spans, as reflected by PIER.
Upweighting POI tokens via WCE yields only marginal improvements over CE on both datasets (e.g., \texttt{cmn-eng} PIER 17.25$\rightarrow$16.68; \texttt{vie-eng} PIER 21.95$\rightarrow$21.18), indicating that emphasizing POIs alone does not reliably disambiguate cross-lingual confusions.
A sequence-level baseline (MWER) improves both WER and PIER relative to CE/WCE, but remains worse than our best contrastive setting on both corpora (e.g., \texttt{cmn-eng} PIER 16.41 vs.\ 15.10; \texttt{vie-eng} PIER 20.84 vs.\ 18.74).

\textbf{Contrastive learning (CL) with near-miss (NM) negatives consistently improves performance.}
CE+CL with $N$-best NMs already beats CE on both datasets, improving \texttt{cmn-eng} WER/PIER (16.67/17.25 $\rightarrow$ 15.64/16.21) and \texttt{vie-eng} WER/PIER (24.72/21.95 $\rightarrow$ 23.16/20.11); adding POI reweighting strengthens this further, and our full tri-level filtered setting achieves the best results (14.06/15.10 and 21.87/18.74). These results support our central hypothesis: CS errors are concentrated at POIs, and learning to rank the reference above acoustically plausible alternatives provides a more informative signal than simply increasing loss weight on POI tokens.

\textbf{Effect of NM source: $N$-best vs.\ $N$-best+LLM.}
Table~\ref{tab:ablations} isolates how candidate sources affect downstream performance.
Under \textit{No filter}, moving from $N$-best to $N$-best+LLM greatly increases the number of available NMs (from 1.40/1.22 to 6.00/6.00 NM/utt for \texttt{cmn-eng}/\texttt{vie-eng}).
However, the impact without gating is \emph{not} uniformly positive: on \texttt{cmn-eng}, LLM expansion improves PIER (15.72$\rightarrow$15.28) but slightly worsens WER (14.93$\rightarrow$15.06), while on \texttt{vie-eng} it improves WER (22.86$\rightarrow$22.19) but degrades PIER (19.10$\rightarrow$19.65).
This mixed behavior suggests that LLM expansion does inject novel confusions beyond beam-search errors, but it can also introduce POI replacements that are not consistently aligned with the audio or with the intended ``hard-but-plausible'' error profile needed for effective contrastive learning.
Importantly, Table~\ref{tab:ablations} shows that \emph{more} negatives are not automatically better: the largest pool (6.0 NM/utt) is not the best-performing configuration, motivating careful selection and filtering.

\textbf{Why tri-level gating is necessary.}
The lower block of Table~\ref{tab:ablations} holds the candidate source fixed to $N$-best+LLM and varies the gating strategy.
Acoustic-only gating retains many candidates (5.75/4.93 NM/utt) yet yields suboptimal WER/PIER on both datasets, consistent with the gate being too permissive.
Adding a phoneme constraint (Ac.+Ph.) improves \texttt{cmn-eng} WER substantially (15.55$\rightarrow$14.50) but does not consistently improve PIER, and it remains weaker than the full pipeline.
Conversely, adding only the text constraint (Ac.+Text) yields strong \texttt{cmn-eng} WER/PIER (14.12/14.69) but fails to transfer to \texttt{vie-eng}, where WER remains high (24.03) and PIER degrades (19.71).
These outcomes highlight that a single constraint captures only one aspect of NM quality: phoneme proximity improves acoustic plausibility, while text distance enforces hardness; either alone can be brittle across languages and scripts.
The full tri-level gate (Ac.+Ph.+Text) combines both constraints and provides the strongest overall WER/PIER trade-off across \texttt{cmn-eng} and \texttt{vie-eng}, while selecting a \emph{smaller} but higher-quality NM set ($\sim$3.8 NM/utt).
Together with Table~\ref{tab:main_results}, this supports the core design of CS-NMG: LLM-expanded near-misses become reliably useful training signal only after jointly enforcing acoustic plausibility, phonetic similarity, and sufficient textual deviation, yielding consistent gains on both global WER and the POI-focused PIER.

\section{Conclusion \& Future work}
We introduced a POI-aware contrastive training framework for code-switching ASR that targets errors concentrated on embedded-language spans and switch-boundary neighborhoods. Our CS-NMG pipeline generates POI-local near-misses by seeding candidates from the ASR $N$-best list, expanding POI replacements with a LLM, and selecting hard-but-plausible negatives via a tri-level gate (acoustic, phoneme, and text). Fine-tuning Whisper-small (LoRA) with a POI-weighted CE anchor plus InfoNCE-style ranking consistently improves both WER and the POI-focused PIER on CS-FLEURS (\texttt{cmn-eng}) and ViMedCSS (\texttt{vie-eng}), without adding auxiliary modules at inference time. 

\textbf{Limitations:} Our near-miss expansion relies on an external LLM API and prompt design, which adds offline cost and can affect reproducibility. Our evaluation is performed on two language pairs and a single model backbone. Future work will explore fully open-source candidate expansion, broader language coverage, and more adaptive gating or negative mining under additional ASR backbones and decoding settings.

\section{Acknowledgments}
This work was conducted as part of the Cross-College project \textbf{\textit{Robust Vietnamese--English Clinical and Educational Medical Translation}} (Project ID: VUNI.2324.CC06), a collaborative initiative between the College of Engineering \& Computer Science and the College of Health Sciences at VinUniversity. 

Tung X. Nguyen, Hieu Minh Truong, Giang Son Nguyen, and Dung D. Le acknowledge partial funding from the Center for AI Research at VinUniversity. Nhu Vo gratefully acknowledges financial support from the Vingroup Scholarship.

\section{Generative AI Use Disclosure}
The authors used generative AI tools only for minor language editing and to improve readability. These tools were not used to generate any scientific content, experimental results, data analyses, or conclusions.

\bibliographystyle{IEEEtran}
\bibliography{mybib}

\end{document}